\pdfoutput=1
\documentclass[11pt]{article}
\usepackage{ACL2023}

\usepackage{amsmath,amsfonts,bm}

\def\eqref#1{equation~\ref{#1}}

\def\1{\bm{1}}

\DeclareMathAlphabet{\mathsfit}{\encodingdefault}{\sfdefault}{m}{sl}
\SetMathAlphabet{\mathsfit}{bold}{\encodingdefault}{\sfdefault}{bx}{n}

\usepackage{times}
\usepackage{latexsym}

\usepackage[T1]{fontenc}

\usepackage[utf8]{inputenc}

\usepackage{microtype}

\usepackage{inconsolata}
\usepackage{hyperref}
\usepackage{url}

\usepackage{graphicx}
\usepackage{url}
\usepackage{booktabs}
\usepackage{multirow}
\usepackage{multicol}

\usepackage{colortbl}
\usepackage{xcolor}

\usepackage{xspace}

\usepackage{pgfplots}
\usepgfplotslibrary{groupplots}
\usepackage{hyperref}
\pgfplotsset{compat=1.18}
\usepackage{filecontents}
\usepackage{wrapfig}
\usepackage{tcolorbox}

\usepackage{gensymb}

\newcommand{\Modelsp}{\textsc{FACT-Bench} }
\newcommand{\Model}{\textsc{FACT-Bench}}
\newcommand{\Premium}{\textsc{Premium2k} }
\newcommand{\Premiumns}{\textsc{Premium2k}}
\newcommand{\vspacesection}{\vspace{-0.5em}}

\title{Towards a Holistic Evaluation of LLMs on Factual Knowledge Recall}

\author{Jiaqing Yuan\textsuperscript{\rm 1}\thanks{Work conducted during an internship at Amazon}, Lin Pan\textsuperscript{\rm 2}\thanks{Corresponding author}, Chung-Wei Hang\textsuperscript{\rm 2}, Jiang Guo\textsuperscript{\rm 2}, Jiarong Jiang\textsuperscript{\rm 2}, Bonan Min\textsuperscript{\rm 2},\\{\bf  Patrick Ng\textsuperscript{\rm 2}, Zhiguo Wang\textsuperscript{\rm 2}}\\
\textsuperscript{\rm 1}North Carolina State University\quad\textsuperscript{\rm 2}AWS AI Labs \\
\texttt{\footnotesize jyuan23@ncsu.edu} \\
\texttt{\footnotesize\{linpan,cwhang,gujiang,jiarongj,bonanmin,patricng,zhiguow\}@amazon.com} \\
}

\begin{document}

\maketitle

\begin{abstract}
Large language models (LLMs) have shown remarkable performance on a variety of NLP tasks, and are being rapidly adopted in a wide range of use cases. It is therefore of vital importance to holistically evaluate the factuality of their generated outputs, as hallucinations remain a challenging issue. 

In this work, we focus on assessing LLMs' ability to recall factual knowledge learned from pretraining, and the factors that affect this ability. To that end, we construct \Model, a representative benchmark covering 20 domains, 134 property types, 3 answer types, and different knowledge popularity levels. We benchmark 31 models from 10 model families and provide a holistic assessment of their strengths and weaknesses. 
We observe that instruction-tuning hurts knowledge recall, as pretraining-only models consistently outperform their instruction-tuned counterparts, and positive effects of model scaling, as larger models outperform smaller ones for all model families. However, the best performance from GPT-4 still represents a large gap with the upper-bound. We additionally study the role of in-context exemplars using counterfactual demonstrations, which lead to significant degradation of factual knowledge recall for large models. By further decoupling model known and unknown knowledge, we find the degradation is attributed to exemplars that contradict a model's known knowledge, as well as the number of such exemplars. Lastly, we fine-tune LLaMA-7B in different settings of known and unknown knowledge. In particular, fine-tuning on a model's known knowledge is beneficial, and consistently outperforms fine-tuning on unknown and mixed knowledge. We will make our benchmark publicly available. 
\end{abstract}

\vspacesection
\section{Introduction}
\vspacesection
Recent advancements of large language models (LLMs), exemplified by ChatGPT\footnote{\url{https://platform.openai.com/docs/models}}, GPT-4 \citep{openai2023gpt4}, are leading to their widespread adoption in various domains. Despite their remarkable performance on NLP tasks, they are still plagued by the issue of hallucinations \citep{hallucination}. Therefore, it is important to conduct holistic assessments to learn how well LLMs capture factual knowledge and what are the factors that affect their ability to recall knowledge learned from pretraining. Previous factuality benchmarks created from knowledge bases \citep{mallen-etal-2023-trust, yu2023kola} focus on a few domains and property types, and questions are created from templates with limited patterns \citep{sun2023headtotail}. Evaluation of LLMs on these benchmarks reveal a large gap from mastery of factual knowledge. However, it is unclear whether such gap is caused by design challenges, such as ambiguity of the questions and presence of multiple plausible answers, which could lead to biased results. 

In this work, we introduce \Model, a comprehensive factuality benchmark consisting of 20K question-answer (QA) pairs and featuring four characteristics: (1) \textit{Simplicity}: we create simple questions from Wikidata triplets (subject, property, object) using Claude \footnote{\url{https://www.anthropic.com/index/introducing-claude}. Specifically, we use \texttt{claude-v1.3-100k} to generate questions.}, to elicit knowledge from LLMs. (2) \textit{Validity}: To make sure the answers are grounded, we select triplets whose subject has a Wikipedia article and whose object also appears in the same article. (3) \textit{Diversity}: \Modelsp covers 20 domains, 134 property types, and 3 answer types (entities, dates and numbers). (4) \textit{Specificity}: we manually select property types that are highly likely to yield unique answers and perform prompt engineering to generate specific questions.

We benchmark 31 models across 10 model families on \Model. Our results reveal that instruction-tuning hurts knowledge recall, as pretraining-only models consistently outperform their instruction-tuned counterparts. We observe positive effects of model scaling --- for all model families, larger models outperform smaller ones across all metrics. However, the best performance from GPT-4 still represents a large gap with the upper-bound. To identify where the gap lies, we conduct evaluation from multiple perspectives and find that LLMs struggle with long-tail entities and certain property types, consistent with the findings in \cite{mallen-etal-2023-trust} and \cite{sun2023headtotail}. 

In addition, we perform counterfactual in-context learning (ICL) experiments to examine the role of in-context exemplars. Our results indicate that counterfactual exemplars lead to significant degradation of factual knowledge recall for large models. By further decoupling model known and unknown knowledge, we find the degradation is attributed to exemplars that contradict a model's \textit{known} knowledge, as well as the number of such exemplars. Lastly, we fine-tune LLaMA-7B in different settings of \textit{known} and \textit{unknown} knowledge. In particular, fine-tuning on knowledge that is \textit{known} to the model is beneficial, and consistently outperforms fine-tuning on knowledge that is \textit{unknown}, which empirically verifies the hypothesis in \citet{schulman-berkeley-talk} that fine-tuning on unknown knowledge teaches the model to hallucinate. 

Our contributions include: (1) A comprehensive benchmark to evaluate LLMs' ability to recall factual knowledge learned from pretraining. (2) Holistic assessment of the strengths and weaknesses of 31 LLMs, and the factors that affect their recall of factual knowledge. (3) Counterfactual ICL experiments to study the role of in-context exemplars, where we find contradicting a model's known knowledge leads to significant degradation of knowledge recall, as well as the number of such exemplars. (4) Fine-tuning experiments that show the advantage of using known knowledge over mixed and unknown knowledge.

\vspacesection
\section{FACT-Bench}
\vspacesection
\label{sec:FACT-Bench}
\subsection{Dataset Construction}
We formulate the factuality evaluation task as closed-book question answering \citep{roberts-etal-2020-much}, where a question is fed to the model without any context, and the model needs to leverage its parametric knowledge to answer the question. As simple as the setup is, we identify four challenges: (1) How to make the questions simple enough so that it solely requires knowledge recall rather than complex reasoning or multi-source information? (2) What types of questions are \textit{fair} to ask? It is unfair to query knowledge that does not exist in the pretraining data of all LLMs. (3) How to make the questions diverse and representative? (4) How to make the question specific enough so that the answer is unique and grounded in some knowledge source?
We address these challenges from the following four aspects.

\textbf{Simplicity.} Although LLMs have shown remarkable performance for solving composite questions \cite{cot}; \cite{leasttomost}, we aim to decouple the ability to reason and to recall factual knowledge. Therefore, we focus on a simple QA setting to elicit knowledge from LLMs and build up the questions based on sampled Wikidata triplets\footnote{We use the dump from \url{https://dumps.wikimedia.org/wikidatawiki/20230601/}.}. The knowledge in Wikidata is in the format of (subject, property, object) triplets, where a simple question can be asked for the property of the subject, and the answer would be the object. 

\textbf{Validity.} To benchmark performance across various models, we take steps to make sure questions in \Modelsp are \textit{answerable} from their pretraining corpora. Although the exact pretraining corpora are not disclosed for some LLMs, it is reasonable to assume that they have all been pretrained on Wikipedia articles. Therefore, we normalize the main content of the Wikipedia page\footnote{We use the \texttt{20220301.en} subset from the Hugging Face datasets library: \url{https://huggingface.co/datasets/wikipedia}.} for the subjects, and only select triplets whose objects also appear in the same Wikipedia page. 

\textbf{Diversity.} We diversify \Modelsp from five aspects: (1) \textit{Multi-domain}. We leverage the knowledge domain categories from Freebase \citep{freebase-2008} and select triplets whose subject has a Wikipedia article page, as well as a Freebase ID. We manually aggregate the 99 top-level domains from Freebase into 20 general domains, such as finance, travel, and literature. (2) \textit{Multi-answer-type}. Unlike previous work, we not only include questions with textual answers, but also dates and numbers. (3) \textit{Multi-property-type}. We manually select a total of 134 diverse properties, which is much more comprehensive than previous benchmarks. The full list of property types by answer type can be found in Appendix~\ref{sec:all properties}. (4) \textit{Multi-knowledge-popularity}. Following previous work \citep{mallen-etal-2023-trust}, we use the view count of subject Wikipedia article from the whole year of 2021 to approximate the popularity of knowledge and sample triplets from the top-25\% and bottom-25\% most popular triplets sets within each domain. (5) \textit{Diverse questions}. Previous benchmarks typically use templates to construct questions from triplets, whereas we leverage a LLM to generate syntactically rich questions. 

\textbf{Specificity.} A challenging issue for the open-domain QA task is that multiple plausible answers may exist for certain questions. We tackle this challenge from two levels. First, select \textit{proper} triplets. For example, the triplet [\textit{\"{O}rjan Sandred, student of, Sven-David Sandstr\"{o}m}] may not be a good triplet since there could be multiple teachers for everyone, whereas the triplet [\textit{Jacob Viner, doctoral advisor, F. W. Taussig}] is more restricted. We manually select property types that are highly likely to yield unique answers. Second, ask \textit{specific} questions. Given a proper triplet, there could be multiple ways to ask questions. For example, given [\textit{Dan Wickline, place of birth, ``Norwalk, California''}], the question \textit{``where was Dan Wickline born?''} has multiple valid answers such as \textit{Norwalk}, and \textit{California}, even though the place of birth is unique for everyone. The question \textit{``What city and state was Dan Wickline born in?''} is more specific. We test multiple prompts for question generation and select one that works best for us (prompt shown in Table~\ref{appendix:question-generation}). Additionally, we filter out triplets whose subjects contain ``()'' in their Wikipedia titles as ``()'' is used for disambiguation\footnote{\url{https://en.wikipedia.org/wiki/Wikipedia:Article_titles\#Disambiguation}}. We also remove triplets that share the same subject and property. Lastly, for specific numerical answers, we check the number together with the unit. For example, for length, we check for 500 kilometers or 500 km instead of just 500, and for temperature, we check for 98 \degree C instead of just 98.

\subsection{Dataset Statistics and Evaluation Metrics}
\label{question distribution}
We manually select 90 properties with textual answers, 22 properties with date answers, and 22 properties with numerical answers. We randomly sample 1000 triplets from each of the 20 domains, where 500 are from the top-25\% most popular triplets, and 500 from the bottom-25\%. The resulting 20k QA pairs are split into training and evaluation set, with a size of 5K and 15K, respectively. The 5K training set is released to facilitate exemplar sampling for ICL and small-scale finetuning. We keep the distribution consistent for any subset, i.e., there is an equal number of examples from each domain, out of which half comes from the top-25\% and the other half from the bottom-25\%.

For evaluation, we use standard metrics for QA tasks, such as SQuAD \citep{rajpurkar-etal-2016-squad}: \textbf{Exact Match} (EM) and \textbf{F1 score}. For answers that are entities, we collect their aliases from Wikidata as additional ground-truth answers. Dates are normalized in the format of \textit{month, day, year}. In zero-shot experiments, we observe models that have not been instruction-tuned tend to generate verbose answers, which leads to low EM and F1 scores but does not necessarily mean that the prediction is wrong. Therefore, we introduce an additional metric \textbf{Contains}, which simply checks if any of the ground-truth answers appear in the prediction.

\subsection{Dataset Validation}
\label{data validation}
We provide a solid estimation of the upper-bound through human validation to validate that \Modelsp is of high quality from the triplet sampling and question generation efforts. 

Concretely, we sample a 2k subset from the 15k evaluation set while keeping the distribution of questions consistent, and manually check the validity and specificity of the questions by examining supporting evidence from Wikipedia articles. We identify 201 questions from the 2k subset that are either ambiguous or not supported by Wikipedia, and replace them with valid ones. Empirically, the upper-bound is 90\% for the 15k set and 100\% for the 2k subset, which we denote as \Premiumns.

\begin{table}[htb!]
\begin{center}
\scalebox{0.55}{%
\begin{tabular}{l|rr|rr|rr|rr}\toprule
\multirow{2}{*}{\bf Models} & \multicolumn{2}{c|}{\bf zero-shot} & \multicolumn{2}{c|}{\bf 1-shot} & \multicolumn{2}{c|}{\bf 6-shot} & \multicolumn{2}{c}{\bf 10-shot} \\ & {\bf EM} & \rotatebox[origin=c]{90}{\bf Contains} & {\bf EM} & \rotatebox[origin=c]{90}{\bf Contains} & {\bf EM} & \rotatebox[origin=c]{90}{\bf Contains} & {\bf EM} & \rotatebox[origin=c]{90}{\bf Contains} \\\midrule
GPT-4  & 58.60 & 64.65 & 59.85 & 63.20 & 63.35 & 66.45 & 65.90 &  69.15 \\
GPT-3.5-turbo  & 49.75 & 52.60 & 51.25 & 53.70 & 52.65 & 55.80 & 53.55 &  56.40 \\ \midrule  
BLOOM-7.1B  & 03.20 & 20.30 & 18.95 & 19.95 & 17.85 & 19.90 & 18.15 &  19.75 \\
BLOOMZ-7.1B  & 18.00 & 19.45 & 14.05 & 15.35 & 14.40 & 17.05 & 15.20 &  17.70 \\  
LLaMA-7B  & 14.65 & 35.20 & 33.25 & 34.15 & 35.55 & 37.15 & 35.05 &  36.75 \\
LLaMA-13B  & 21.35 & 39.95 & 36.45 & 37.30 & 41.15 & 42.75 & 41.20 &  42.95 \\
LLaMA-33B  & 27.25 & 46.55 & 45.25 & 46.70 & 48.30 & 50.30 & 48.90 &  51.10 \\  
LLaMA-65B  & 35.25 & 49.20 & 47.15 & 48.45 & 52.15 & 53.80 & 52.45 &  54.10 \\
Vicuna-7B-v1.3  & 24.65 & 33.25 & 31.15 & 33.80 & 30.10 & 35.05 & 31.00 &  34.65 \\ 
Vicuna-13B-v1.3  & 32.95 & 35.15 & 36.45 & 37.60 & 38.00 & 41.20 & 38.40 &  41.15 \\
Vicuna-33B-v1.3  & 34.30 & 44.15 & 41.39 & 44.75 & 44.10 & 48.10 & 44.00 &  48.05 \\
OpenLLaMA-7B  & 14.05 & 32.30 & 31.75 & 32.80 & 32.55 & 34.70 & 33.80 &  35.95 \\
OpenLLaMA-13B  & 25.70 & 37.35 & 37.05 & 38.40 & 38.75 & 40.70 & 39.70 &  41.55 \\
FLAN-T5-XXL (11B)  & 20.60 & 21.60 & 20.45 & 21.45 & 21.05 & 22.15 & 20.95 &  22.00 \\  
T0++ (11B)  & 16.05 & 21.25 & 16.75 & 19.95 & 16.80 & 20.00 & 17.05 &  19.85 \\
UL2 (20B)  & 03.40 & 23.55 & 23.50 & 24.40 & 24.15 & 25.75 & 23.50 &  25.00 \\
FLAN-UL2 (20B)  & 24.05 & 25.20 & 24.10 & 25.25 & 24.10 & 25.30 & 23.90 &  24.95 \\
Falcon-7B  & 23.60 & 30.05 & 30.25 & 31.90 & 30.70 & 32.60 & 30.45 &  32.25 \\
Falcon-7B-instruct  & 10.85 & 25.10 & 21.75 & 24.60 & 22.45 & 25.45 & 22.45 &  25.20 \\  
Falcon-40B  & 26.55 & 30.90 & 39.10 & 40.50 & 42.05 & 43.60 & 42.25 &  43.80 \\
Falcon-40B-instruct  & 21.95 & 40.25 & 38.85 & 40.75 & 40.40 & 42.20 & 40.00 &  41.85 \\
Falcon-180B  & 44.90 & 47.45 & 49.25 & 50.60 & 53.55 & 55.05 & 53.45 &  55.00 \\
Falcon-180B-chat  & 39.95 & 47.10 & 47.00 & 49.30 & 49.05 & 51.50 & 49.30 &  51.60 \\
MPT-7B  & 03.45 & 30.35 & 28.85 & 29.85 & 29.75 & 31.15 & 30.45 &  31.55 \\  
MPT-7B-instruct  & 03.55 & 30.40 & 21.55 & 29.25 & 26.35 & 29.30 & 27.85 &  29.60 \\
MPT-30B  & 25.30 & 35.00 & 34.35 & 35.55 & 35.80 & 37.55 & 36.05 &  37.75 \\
MPT-30B-instruct  & 19.05 & 33.50 & 28.80 & 31.20 & 31.00 & 33.50 & 31.50 &  33.85 \\  
Pythia-6.9B  & 11.00 & 13.15 & 21.20 & 22.45 & 21.85 & 23.05 & 21.70 &  23.25 \\
Pythia-12B  & 15.25 & 22.00 & 22.75 & 23.70 & 22.95 & 24.35 & 23.20 &  24.65 \\
Mistral-7B  & 28.45 & 29.25 & 38.90 & 39.80 & 40.45 & 41.85 & 40.75 &  42.60 \\
Mistral-7B-instruct  & 26.00 & 29.30 & 26.80 & 30.05 & 26.80 & 30.35 & 27.20 &  30.75 \\
\bottomrule
\end{tabular}%
}%
\end{center}
\caption{Benchmarking results on \Premiumns.}
\label{table: all results}
\end{table}
\begin{figure*}[htb]
    \centering
    \centering
\begin{tikzpicture}\tikzstyle{every node}=[font=\tiny]
\begin{axis}[
    ybar=0pt,
    width=1.0\textwidth,
    height=3.8cm,
    xtick=data,
    xticklabels from table={popularity.dat}{Models},
    x tick label style={rotate=45,anchor=east,font=\tiny},
     ylabel={Exact Match},
    ymin=0,
    ymax=90,
    bar width=3pt,
    enlarge x limits=0.04,
    legend entries={Top-25\%,Bottom-25\%},
    legend style={at={(0.5,0.7)},anchor=south,legend columns=3},
]
\foreach \colindex in {1,2} {
    \addplot table [y index=\colindex, x expr=\coordindex] {popularity.dat};
}
\end{axis}
\end{tikzpicture}
    \vspace{-6mm}
    \caption{10-shot EM by knowledge popularity. Knowledge popularity is a strong predictor of knowledge recall. LLMs struggle with long-tail entities (Bottom-25\%) as shown by the large gap with popular entities (Top-25\%).}
    \label{fig:popularity-top-bottom}
\end{figure*}
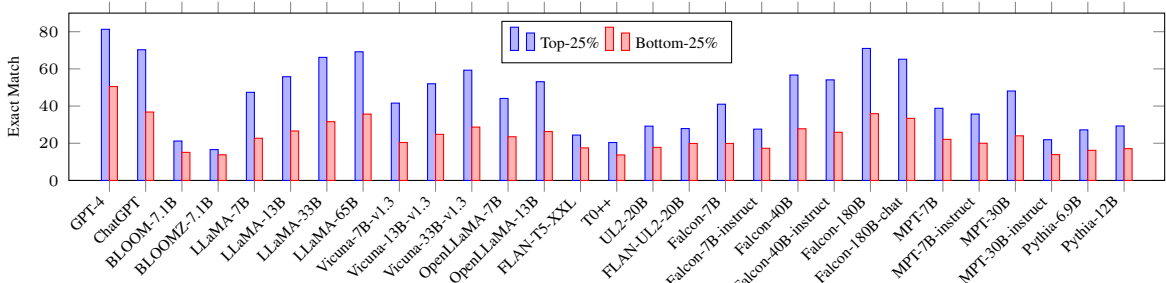
\begin{figure*}[htb]
\begin{center}
\pgfplotsset{
   /pgfplots/bar  cycle  list/.style={/pgfplots/cycle  list={%
        {violet,fill=violet!30!white,mark=none},%
        {red,fill=red!30!white,mark=none},%
        {brown,fill=brown!30!white,mark=none},%
        {green,fill=green!30!white,mark=none},%
        {blue,fill=blue!30!white,mark=none},%
        {gray,fill=gray!30!white,mark=none},%
     }
   },
}

\scalebox{0.8}{
    \begin{tikzpicture}\tikzstyle{every node}=[font=\tiny]
        \begin{axis}[
            ybar=0pt,  
            width=1.3\textwidth,
            height=4.5cm,
            ymin=0,
            ymax=100,
            ylabel={Exact Match},
            symbolic x coords={
                all,country,located in admin territorial entity,place of birth,inception,
                country of citizenship,date of birth,occupation,place of death,sport,manufacturer,
                taxon rank,educated at,date of death,country of origin,headquarters location,part of,
                parent taxon
            },
            xtick=data,
            x tick label style={rotate=25,anchor=east},
            legend style={at={(0.5,-0.25)}, anchor=north,legend columns=-1},
            bar width=3pt, 
            enlarge x limits={abs=15pt}, 
            xtick distance=1.5, 
        ]
        
        \addplot coordinates {
            (all,67.41) (country,95.38) (located in admin territorial entity,61.11)
            (place of birth,46.77) (inception,59.18) (country of citizenship,91.8) (date of birth,53.98)
            (occupation,73.33) (place of death,59.57) (sport,100) (manufacturer,60) (taxon rank,100)
            (educated at,55.26) (date of death,43.06) (country of origin,92.11) (headquarters location,65.52)
            (part of,66.67) (parent taxon,55.56)
        };
        
        \addplot coordinates {
            (all,59) (country,93.85) (located in admin territorial entity,55.56)
            (place of birth,38.71) (inception,54.08) (country of citizenship,91.8) (date of birth,44.25)
            (occupation,53.33) (place of death,46.81) (sport,64.71) (manufacturer,40) (taxon rank,100)
            (educated at,28.95) (date of death,27.78) (country of origin,86.84) (headquarters location,62.07)
            (part of,55.56) (parent taxon,50)
        };
        

        \addplot coordinates {
            (all,37.85)
            (country,76.15)
            (located in admin territorial entity,41.67)
            (place of birth,24.19)
            (inception,25.51)
            (country of citizenship,80.33)
            (date of birth,12.39)
            (occupation,40.0)
            (place of death,27.66)
            (sport,64.71)
            (manufacturer,60)
            (taxon rank,11.11)
            (educated at,15.79)
            (date of death,5.56)
            (country of origin,63.16)
            (headquarters location,31.03)
            (part of,29.63)
            (parent taxon,22.22)
        };

        \addplot coordinates {
            (all,44.51)
            (country,83.85)
            (located in admin territorial entity,44.44)
            (place of birth,19.35)
            (inception,36.73)
            (country of citizenship,80.33)
            (date of birth,17.7)
            (occupation,46.67)
            (place of death,34.04)
            (sport,76.47)
            (manufacturer,50)
            (taxon rank,100.0)
            (educated at,21.05)
            (date of death,9.72)
            (country of origin,73.68)
            (headquarters location,41.38)
            (part of,40.74)
            (parent taxon,38.89)
        };
        
        \addplot coordinates {
            (all,52.92) (country,88.46) (located in admin territorial entity,47.22)
            (place of birth,27.42) (inception,44.9) (country of citizenship,88.52) (date of birth,34.51)
            (occupation,60) (place of death,36.17) (sport,76.47) (manufacturer,70) (taxon rank,100)
            (educated at,28.95) (date of death,23.61) (country of origin,73.68) (headquarters location,55.17)
            (part of,51.85) (parent taxon,50)
        };

        \addplot coordinates {
            (all,55.37)
            (country,91.54)
            (located in admin territorial entity,50.0)
            (place of birth,33.87)
            (inception,50.0)
            (country of citizenship,85.25)
            (date of birth,35.4)
            (occupation,53.33)
            (place of death,38.3)
            (sport,82.35)
            (manufacturer,90)
            (taxon rank,100.0)
            (educated at,34.21)
            (date of death,18.06)
            (country of origin,84.21)
            (headquarters location,55.17)
            (part of,55.56)
            (parent taxon,55.56)
        };

        
        \legend{GPT-4, ChatGPT, LLaMA-7B, LLaMA-13B, LLaMA-33B, LLaMA-65B}
        \end{axis}
    \end{tikzpicture}}
\end{center}
\vspace{-5mm}
\caption{10-shot EM by property type. LLMs do well on certain property types, such as country-related properties, while struggle on other property types, such as date-related properties. Due to space, we show results for GPT and LLaMA models, and the most common property types from the full set of 134 property types.}
\label{fig:by-property-type}
\end{figure*}
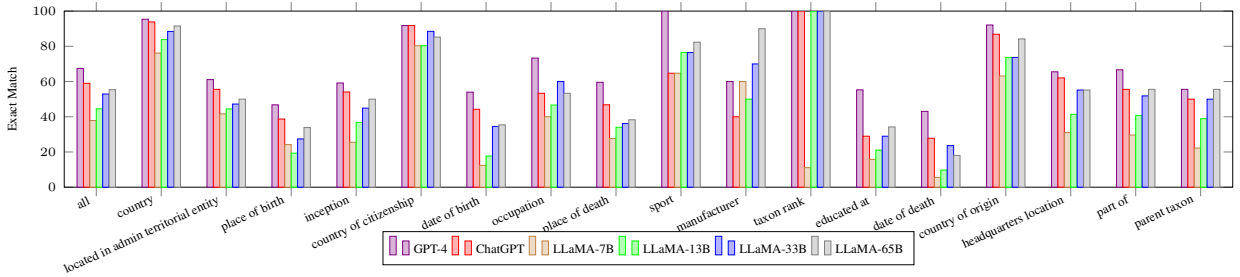

\vspacesection
\section{Benchmarking LLMs}
\vspacesection
\label{sec:Benchmarking LLMs}

\subsection{Experimental Setup}

We consider LLMs with different model architectures, sizes, pretraining-only/instruction-tuning, and conduct zero-shot and few-shot ICL experiments. Specifically, we benchmark GPT-4, GPT-3.5-turbo\footnote{We access the APIs of OpenAI models from the week of July 3rd to that of July 17th, 2023}, BLOOM/BLOOMZ (7B) \citep{workshop2023bloom}, LLaMA (7B, 13B, 33B, 65B) \citep{touvron2023llama}, Vicuna (7B, 13B, 33B) \citep{vicuna2023}, OpenLLaMA (7B, 13B) \citep{openlm2023openllama}, FLAN-T5-XXL (11B) \citep{chung2022scaling}, T0++ (11B) \citep{sanh2021multitask}, UL2/FLAN-UL2 (20B) \citep{tay2023ul2}, Falcon/Falcon-instruct (7B, 40B, 180B) \citep{falcon40B}, MPT/MPT-instruct (7B, 30B) \citep{MosaicML2023Introducing}, Pythia (6.9B, 12B) \citep{biderman2023pythia}, and Mistral/Mistral-instruct (7B) \citep{jiang2023mistral}. For all LLMs, we use the same prompts shown in Table~\ref{appendix:zero-shot} and~\ref{appendix:10-shot}. The exemplars in the few-shot experiments are shared across models and are randomly sampled from the training set, considering coverage for all 3 answer types (entities, dates and numbers). All our experiments are conducted on the \Premium subset to reduce the cost of running LLMs\footnote{For the full 15k evaluation set, we provide zero-shot and 10-shot results in Appendix~\ref{subsec:15k eval} for reference.}.

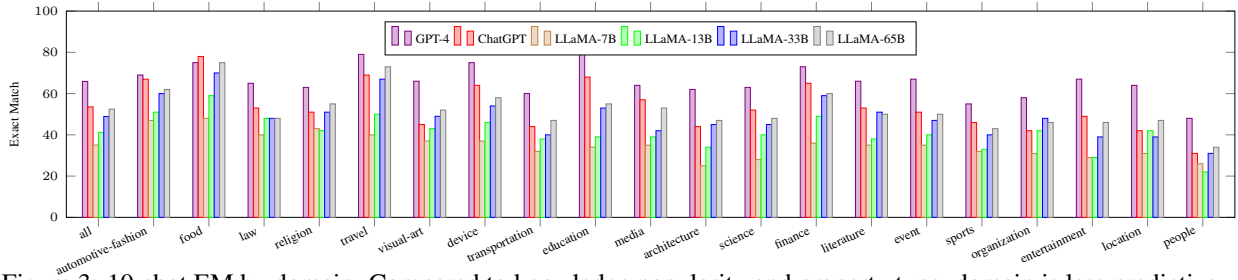
\begin{figure*}
\begin{center}
\pgfplotsset{
   /pgfplots/bar  cycle  list/.style={/pgfplots/cycle  list={%
        {violet,fill=violet!30!white,mark=none},%
        {red,fill=red!30!white,mark=none},%
        {brown,fill=brown!30!white,mark=none},%
        {green,fill=green!30!white,mark=none},%
        {blue,fill=blue!30!white,mark=none},%
        {gray,fill=gray!30!white,mark=none},%
     }
   },
}

\scalebox{0.8}{%
\begin{tikzpicture}\tikzstyle{every node}=[font=\tiny]
    \begin{axis}[
        ybar=0pt,  
        width=1.3\textwidth,
        height=5cm,
        ymin=0,
        ymax=100,
        ylabel={Exact Match},
        symbolic x coords={
            all,automotive-fashion,food,law,religion,travel,visual-art,device,transportation,
            education,media,architecture,science,finance,literature,event,sports,organization,
            entertainment,location,people
        },
        xtick=data,
        x tick label style={rotate=25,anchor=east},
        legend style={at={(0.5,0.95)}, anchor=north,legend columns=-1},
        bar width=2.5pt, 
        enlarge x limits={abs=15pt}, 
        xtick distance=1.5, 
    ]
    
    \addplot coordinates {
        (all,65.9) (automotive-fashion,69) (food,75) (law,65) (religion,63) (travel,79)
        (visual-art,66) (device,75) (transportation,60) (education,79) (media,64) (architecture,62)
        (science,63) (finance,73) (literature,66) (event,67) (sports,55) (organization,58)
        (entertainment,67) (location,64) (people,48)
    };
    
    \addplot coordinates {
        (all,53.55) (automotive-fashion,67) (food,78) (law,53) (religion,51) (travel,69)
        (visual-art,45) (device,64) (transportation,44) (education,68) (media,57) (architecture,44)
        (science,52) (finance,65) (literature,53) (event,51) (sports,46) (organization,42)
        (entertainment,49) (location,42) (people,31)
    };
    

    \addplot coordinates {
        (all,35.05)
        (automotive-fashion,47)
        (food,48)
        (law,40)
        (religion,43)
        (travel,40)
        (visual-art,37)
        (device,37)
        (transportation,32)
        (education,34)
        (media,35)
        (architecture,25)
        (science,28)
        (finance,36)
        (literature,35)
        (event,35)
        (sports,32)
        (organization,31)
        (entertainment,29)
        (location,31)
        (people,26)
    };

    \addplot coordinates {
        (all,41.2)
        (automotive-fashion,51)
        (food,59)
        (law,48)
        (religion,42)
        (travel,50)
        (visual-art,43)
        (device,46)
        (transportation,38)
        (education,39)
        (media,39)
        (architecture,34)
        (science,40)
        (finance,49)
        (literature,38)
        (event,40)
        (sports,33)
        (organization,42)
        (entertainment,29)
        (location,42)
        (people,22)
    };

    \addplot coordinates {
        (all,48.9)
        (automotive-fashion,60)
        (food,70)
        (law,48)
        (religion,51)
        (travel,67)
        (visual-art,49)
        (device,54)
        (transportation,40)
        (education,53)
        (media,42)
        (architecture,45)
        (science,45)
        (finance,59)
        (literature,51)
        (event,47)
        (sports,40)
        (organization,48)
        (entertainment,39)
        (location,39)
        (people,31)
    };

    \addplot coordinates {
        (all,52.45)
        (automotive-fashion,62)
        (food,75)
        (law,48)
        (religion,55)
        (travel,73)
        (visual-art,52)
        (device,58)
        (transportation,47)
        (education,55)
        (media,53)
        (architecture,47)
        (science,48)
        (finance,60)
        (literature,50)
        (event,50)
        (sports,43)
        (organization,46)
        (entertainment,46)
        (location,47)
        (people,34)
    };

    
    \legend{GPT-4, ChatGPT, LLaMA-7B, LLaMA-13B, LLaMA-33B, LLaMA-65B}
    \end{axis}
\end{tikzpicture}}
\end{center}
\vspace{-17mm}
\caption{10-shot EM by domain. Compared to knowledge popularity and property type, domain is less predictive of knowledge recall as model performances across different domains are more flat. Due to space, we show results for GPT and LLaMA models.}
\label{fig:by-domain}
\end{figure*}

\begin{figure}
\begin{center}
\pgfplotsset{
   /pgfplots/bar  cycle  list/.style={/pgfplots/cycle  list={%
        {violet,fill=violet!30!white,mark=none},%
        {red,fill=red!30!white,mark=none},%
        {brown,fill=brown!30!white,mark=none},%
        {green,fill=green!30!white,mark=none},%
        {blue,fill=blue!30!white,mark=none},%
        {gray,fill=gray!30!white,mark=none},%
     }
   },
}

\scalebox{1}{
    \begin{tikzpicture}\tikzstyle{every node}=[font=\tiny]
        \begin{axis}[
            ybar=0pt,  
            width=.5\textwidth,
            height=3.5cm,
            ymin=0,
            ymax=80,
            ylabel={Exact Match},
            symbolic x coords={
                all,entity,date,number
            },
            xtick=data,
            x tick label style={rotate=25,anchor=east},
            legend style={at={(0.5,-0.25)}, anchor=north,legend columns=3},
            bar width=4pt, 
            enlarge x limits={.25}, 
        ]
        
        \addplot coordinates {
            (all,65.9) (entity,70.05) (date,56.84) (number,54.35)
        };
        
        \addplot coordinates {
            (all,53.55) (entity,58.1) (date,44.12) (number,38.04)
        };
        

        \addplot coordinates {
            (all,35.05) (entity,42.76) (date,17.15) (number,19.57)
        };

        \addplot coordinates {
            (all,41.2) (entity,48.24) (date,24.47) (number,29.35)
        };

        \addplot coordinates {
            (all,48.9) (entity,54.79) (date,35.26) (number,36.96)
        };

        \addplot coordinates {
            (all,52.45) (entity,58.96) (date,37.96) (number,35.87)
        };

        
        \legend{GPT-4, ChatGPT, LLaMA-7B, LLaMA-13B, LLaMA-33B, LLaMA-65B}
        \end{axis}
    \end{tikzpicture}}
\end{center}
\vspace{-5mm}
\caption{10-shot EM by answer type. LLMs are less capable on date and numerical knowledge. Due to space, we show results for GPT and LLaMA models.}
\label{fig:by-answer-type}
\end{figure}
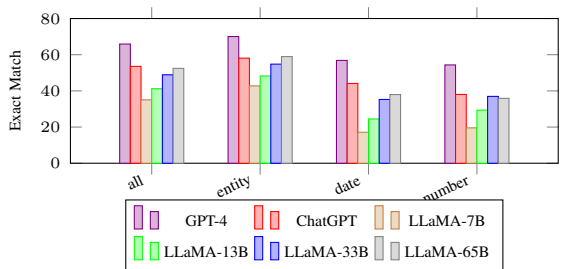

\subsection{Results}
Benchmarking results are presented in Table~\ref{table: all results}\footnote{Full results including F1 scores can be found in Appendix~\ref{subsec:full premium2k}.}.  

\textbf{Large gap with upper-bound}. GPT-4 outperforms all the other models we consider on our benchmark. However, its performance of 65.9\% EM in the 10-shot setting still represents a large gap with the estimated upper-bound, which shows the challenge of mastering factuality, as well as the potential risks of using LLMs in certain tasks.

\textbf{Positive effect of model scaling}.
Overall, we observe positive effects of model scaling. For all model families (i.e., LLaMA, Falcon, and MPT), larger model sizes translate to better performances across settings. Closed-source GPT models significantly outperform open-source models with the notable exception of LLaMA-65B, which is competitive with GPT-3.5-turbo in the 10-shot setting.

\textbf{Negative impact of instruction-tuning}. Comparing models in their pretraining-only form and their instruction-tuned counterparts, such as LLaMA/Vicuna, Falcon/Falcon-instruct, and MPT/MPT-instruct in the few-shot setting, all instruction-tuned models display inferior performance for all metrics. In the zero-shot setting, pretraining-only models tend to generate verbose answers, which leads to low EM and F1 scores, but the \textit{Contains} metric reveals that they outperform their instruction-tuned counterparts. This result empirically verifies the hypothesis in \citet{zhou2023lima} that most knowledge of LLMs is learned during pretraining and alignment only helps with output style and format. We hypothesize that the \textit{alignment tax} \citep{instructgpt} from instruction-tuning leads to the performance drop. Overall, the best performance for each model family is achieved by few-shot ICL with the pretraining-only version of the model.

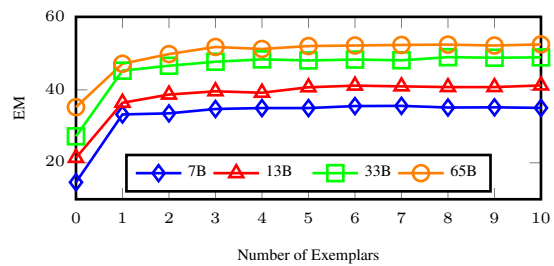
\begin{figure}[htb]
\begin{center}
\begin{tikzpicture}\tikzstyle{every node}=[font=\tiny]
\begin{axis}[
    width=1\linewidth,
    height=4cm,
    xlabel={Number of Exemplars},
    ylabel={EM},
    xmin=0, xmax=10,
    ymin=10, ymax=60,
    line width=1.0pt,
    mark size=3.0pt,
    xtick=data,
    legend entries={7B, 13B, 33B, 65B},
    typeset ticklabels with strut,
    legend columns=4, 
        legend style={
            /tikz/column 4/.style={
                column sep=5pt,
            },
            at={(0.5,0.25)},anchor=north,
        },
    legend cell align=left,
]

\addplot[mark=diamond, blue] coordinates {
    (0, 14.65)
    (1, 33.25)
    (2, 33.55)
    (3, 34.75)
    (4, 35.00)
    (5, 35.00)
    (6, 35.55)
    (7, 35.60)
    (8, 35.15)
    (9, 35.20)
    (10, 35.05)
};

\addplot[mark=triangle, red] coordinates {
    (0, 21.35)
    (1, 36.45)
    (2, 38.7)
    (3, 39.6)
    (4, 39.2)
    (5, 40.7)
    (6, 41.15)
    (7, 40.95)
    (8, 40.75)
    (9, 40.75)
    (10, 41.2)
};

\addplot[mark=square, green] coordinates {
    (0, 27.25)
    (1, 45.25)
    (2, 46.6)
    (3, 47.7)
    (4, 48.35)
    (5, 48.05)
    (6, 48.3)
    (7, 48.1)
    (8, 48.95)
    (9, 48.8)
    (10, 48.9)
};

\addplot[mark=o, orange] coordinates {
    (0, 35.25)
    (1, 47.15)
    (2, 49.8)
    (3, 51.75)
    (4, 51.2)
    (5, 52)
    (6, 52.15)
    (7, 52.3)
    (8, 52.4)
    (9, 52.15)
    (10, 52.45)
};
\end{axis}
\end{tikzpicture}
\end{center}
\vspace{-15mm}
\caption{LLaMA zero-to-10-shot results by EM.}
\label{fig:llama-0-10-shots-em}
\end{figure}

\textbf{Diminishing returns from adding more exemplars}. Going from zero-shot to 1-shot, all open-source models benefit greatly learning from the answer format of the in-context exemplar, which is reflected in their improved EM scores. This is especially the case for pretraining-only models. By the Contains metric, results are mixed. As \textit{k} increases to 6, all models, with the exception of BLOOM and T0++, show improvements over zero-shot and 1-shot. However, going from 6-shot to 10-shot, model performances mostly stay flat, except for GPT-4, improving by 2.7\%. To further validate this, we run zero, 1--10 shots with LLaMA models, and results are shown in Figure~\ref{fig:llama-0-10-shots-em}, where the curves flatten after providing 3--5 exemplars. 


\subsection{Fine-grained Evaluation}
\label{subsec:finegrained-eval}

To gain a better understanding of where the gap with the upper-bound lies, we examine model performances from multiple perspectives. 

\textbf{Knowledge popularity and property type are predictive of knowledge recall}. Figure~\ref{fig:popularity-top-bottom} shows 10-shot performance by knowledge popularity and Figure~\ref{fig:by-property-type} by property type. We observe similar findings in \citet{mallen-etal-2023-trust} that knowledge popularity and property type are strong predictors of knowledge recall. LLMs struggle with long-tail entities (Bottom-25\%) as shown by the large gap with popular entities (Top-25\%). This result suggests that knowledge distribution of the \textit{pretraining data} (if known to the model user) can potentially be leveraged as a predictor for factual knowledge recall. LLMs do well on certain property types, such as country-related properties, while struggle on other property types, such as date-related properties. Further results by answer type (Figure~\ref{fig:by-answer-type}) show that LLMs are less capable on date and numerical knowledge.

\textbf{Domain is less predictive of knowledge recall}. On the other hand, domain is not a strong predictor of model performance as shown in Figure~\ref{fig:by-domain}, where model performances across different domains are more flat compared to knowledge popularity levels and property types.

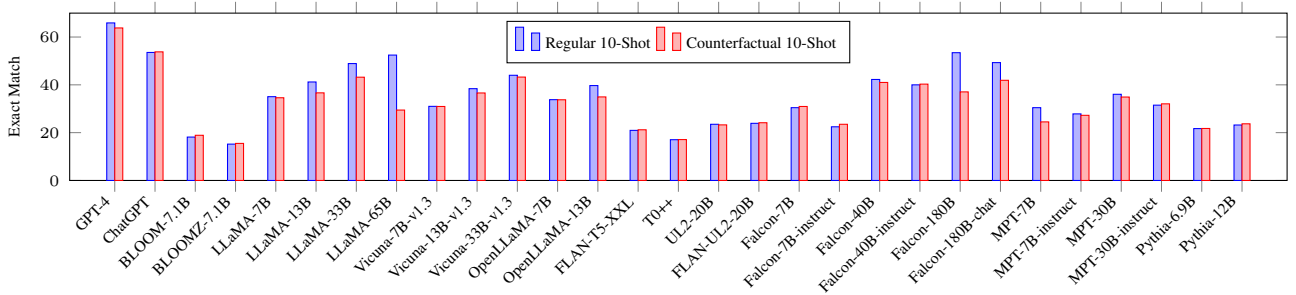
\begin{figure*}[htb]
    \centering
    \centering
\begin{tikzpicture}\tikzstyle{every node}=[font=\tiny]
\begin{axis}[
    ybar=0pt,
    width=1.1\textwidth,
    height=3.8cm,
    xtick=data,
    xticklabels from table={counterfactual_or_not.csv}{Models},
    x tick label style={rotate=45,anchor=east,font=\tiny},
    ylabel={Exact Match},
    ymin=0,
    ymax=70,
    bar width=3pt,
    enlarge x limits=0.04,
    legend entries={Regular 10-Shot,Counterfactual 10-Shot},
    legend style={at={(0.5,0.7)},anchor=south,legend columns=3},
]
\foreach \colindex in {1,2} {
    \addplot table [y index=\colindex, x expr=\coordindex] {counterfactual_or_not.csv};
}
\end{axis}
\end{tikzpicture}
    \vspace{-8mm}
    \caption{Comparison of regular 10-shot and counterfactual 10-shot by Exact Match. LLaMA-65B experiences a major drop with counterfactual exemplars, followed by Falcon-180B and LLaMA-33B, while the performance of smaller models remains flat and unaffected.}
    \label{fig:counterfactual-or-not}
\end{figure*}

\vspacesection
\section{The Role of In-context Exemplars}
\vspacesection
\label{sec:The Role of ICL Demonstration}
Previous work \cite{min-etal-2022-rethinking} suggests that ground-truth labels play an insignificant role for ICL, such that replacing ground-truth labels with random labels on classification and multi-choice tasks only results in marginal loss of accuracy. Compared to classification and multi-choice tasks, the label space of our task is much larger. We design a set of experiments to investigate how \textit{counterfactual} in-context exemplars affect a model’s ability to recall factual knowledge. 

\subsection{Counterfactual ICL}
\textbf{Experimental setup}.
In this set of experiments, we replace the ground-truth answers of our regular 10-shot exemplars with random answers chosen from the 5k training set. We impose an additional constraint that the random answer is chosen within the same property type, denoted as \textbf{shuffle}. For example, we change the ground-truth answer for \textit{``In which military branch did Henry Curtis serve?''} from \textit{``Royal Navy''} to the counterfactual answer \textit{``United States Marine Corps''}. Without prior knowledge required to answer the question, the new input-label pair looks reasonable but is actually not factual.

\textbf{Results}.
Figure~\ref{fig:counterfactual-or-not} shows the results. Notably, LLaMA-65B experiences a major drop from 52.45\% EM (regular 10-shot) to 29.45\%, followed by Falcon-180B from 53.45\% to 37.05\%, and LLaMA-33B from 48.9\% to 43.2\%, while the performance of \textit{smaller} models remains flat and unaffected. 
In addition, we observe that instruction-tuned models are less affected by counterfactual exemplars than their pretraining-only counterparts. For example, compared to LLaMA, Falcon and MPT, the drop is less significant for Vicuna, Falcon-chat, and MPT-instruct models, respectively. 

\subsection{Counterfactual ICL with known and unknown knowledge}
\label{subsec:icl counterfactual known unknown}
Results in the previous section show that counterfactual exemplars lead to significant degradation of factual knowledge recall for large models. However, it is not clear what factors lead to this behavior besides model scale. LLaMA-65B, Falcon-180B and LLaMA-33B are the three most capable open-source models on our benchmark (Table~\ref{table: all results}). Since the 10 in-context exemplars are randomly sampled, it is expected that these three models have more knowledge about the exemplars than the other models. Therefore, we further decouple \textit{known} and \textit{unknown} knowledge of the exemplars to study their role.

\textbf{Experimental setup}.
We conduct controlled experiments on LLaMA models and \texttt{text-davinci-002}\footnote{In \citet{wei2023larger}, experiments using in-context exemplars with flipped labels show that \texttt{text-davinci-002} experiences the largest drop on binary classification tasks. We further include this model in this set of experiments.}. To approximate model known and unknown knowledge, we sample \textit{k} = 32 questions that are correctly answered by each model as \textbf{known} knowledge, and \textit{k} = 32 incorrectly answered as \textbf{unknown} knowledge. We corrupt the exemplars with the same shuffling method as the previous experiment.


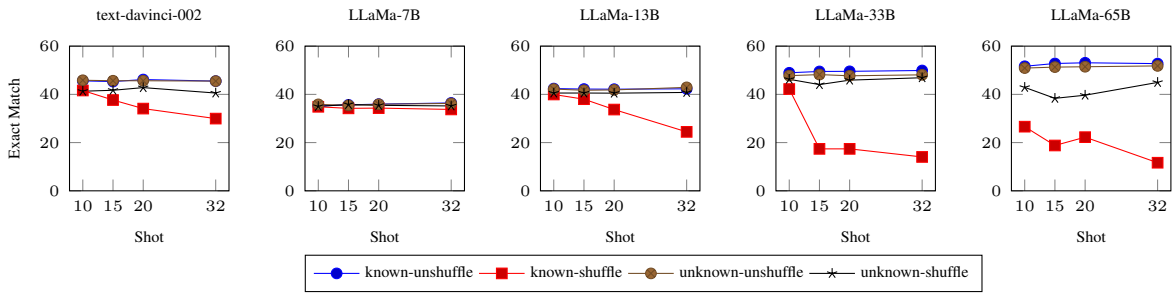
\begin{figure*}[htb]
\begin{center}
\pgfplotsset{
   /pgfplots/bar  cycle  list/.style={/pgfplots/cycle  list={%
        {violet,fill=violet!30!white,mark=none},%
        {red,fill=red!30!white,mark=*},%
        {brown,fill=brown!30!white,mark=square},%
        {green,fill=green!30!white,mark=otimes},%
        {blue,fill=blue!30!white,mark=triangle},%
        {gray,fill=gray!30!white,mark=diamond},%
     }
   },
}
\begin{tikzpicture} \tikzstyle{every node}=[font=\tiny]
\begin{groupplot}[group style={group size=5 by 1},width=.23\textwidth,height=3.5cm]
\nextgroupplot[
    xlabel style={align=center}, xlabel={Shot},
    title={text-davinci-002},
    ylabel={Exact Match},
    ymin=0,
    ymax=60,
    ytick={0,20,40,60},
    xtick={10,15,20,32},
    legend style={at={(3.6,-0.7)},anchor=south,legend columns=4,font=\tiny}
]
\addplot table [x=shot, y=known-unshuffle, col sep=comma] {text-davinci-002.csv};
\addlegendentry{known-unshuffle}
\addplot table [x=shot, y=known-shuffle-by-property, col sep=comma] {text-davinci-002.csv};
\addlegendentry{known-shuffle}
\addplot table [x=shot, y=unknown-unshuffle, col sep=comma] {text-davinci-002.csv};
\addlegendentry{unknown-unshuffle}
\addplot table [x=shot, y=unknown-shuffle-by-property, col sep=comma] {text-davinci-002.csv};
\addlegendentry{unknown-shuffle}

\nextgroupplot[
    xlabel style={align=center}, xlabel={Shot},
    title={LLaMa-7B},
    ymin=0,
    ymax=60,
    ytick={0,20,40,60},
    xtick={10,15,20,32},
]
\addplot table [x=shot, y=known-unshuffle, col sep=comma] {LLaMa-7b.csv};
\addplot table [x=shot, y=known-shuffle-by-property, col sep=comma] {LLaMa-7b.csv};
\addplot table [x=shot, y=unknown-unshuffle, col sep=comma] {LLaMa-7b.csv};
\addplot table [x=shot, y=unknown-shuffle-by-property, col sep=comma] {LLaMa-7b.csv};

\nextgroupplot[
    xlabel style={align=center}, xlabel={Shot},
    title={LLaMa-13B},
    ymin=0,
    ymax=60,
    ytick={0,20,40,60},
    xtick={10,15,20,32},
]
\addplot table [x=shot, y=known-unshuffle, col sep=comma] {LLaMa-13b.csv};
\addplot table [x=shot, y=known-shuffle-by-property, col sep=comma] {LLaMa-13b.csv};
\addplot table [x=shot, y=unknown-unshuffle, col sep=comma] {LLaMa-13b.csv};
\addplot table [x=shot, y=unknown-shuffle-by-property, col sep=comma] {LLaMa-13b.csv};

\nextgroupplot[
    xlabel style={align=center}, xlabel={Shot},
    title={LLaMa-33B},
    ymin=0,
    ymax=60,
    ytick={0,20,40,60},
    xtick={10,15,20,32},
]
\addplot table [x=shot, y=known-unshuffle, col sep=comma] {LLaMa-33b.csv};
\addplot table [x=shot, y=known-shuffle-by-property, col sep=comma] {LLaMa-33b.csv};
\addplot table [x=shot, y=unknown-unshuffle, col sep=comma] {LLaMa-33b.csv};
\addplot table [x=shot, y=unknown-shuffle-by-property, col sep=comma] {LLaMa-33b.csv};

\nextgroupplot[
    xlabel style={align=center}, xlabel={Shot},
    title={LLaMa-65B},
    ymin=0,
    ymax=60,
    ytick={0,20,40,60},
    xtick={10,15,20,32},
]
\addplot table [x=shot, y=known-unshuffle, col sep=comma] {LLaMa-65b.csv};
\addplot table [x=shot, y=known-shuffle-by-property, col sep=comma] {LLaMa-65b.csv};
\addplot table [x=shot, y=unknown-unshuffle, col sep=comma] {LLaMa-65b.csv};
\addplot table [x=shot, y=unknown-shuffle-by-property, col sep=comma] {LLaMa-65b.csv};
\end{groupplot}
\end{tikzpicture}
\end{center}
\vspace{-2mm}
\caption{Counterfactual few-shot with known and unknown knowledge, evaluated by Exact Match. Result shows that the degradation in factual knowledge recall is primarily due to exemplars that contradict models' \textit{known} knowledge, and the number of such exemplars.}
\label{fig:counterfactual-few-shot-contrastive}
\end{figure*}


\textbf{Contradicting LLMs' known knowledge teaches them to lie}. Results are shown in Figure~\ref{fig:counterfactual-few-shot-contrastive}. Comparing \textit{known-shuffle} with \textit{unknown-shuffle} in the 10-shot setting, LLaMA-65B drops from 52.45\% EM (regular 10-shot) to 26.60\% with known-shuffle while the drop with unknown-shuffle is much less significant from 52.45\% EM to 42.90\%. For LLaMA-33B, performance drops from 48.30\% to 42.20\% with known-shuffle, and from 48.30\% to 46.25\% with unknown-shuffle. For the larger text-davinci-002 model, performances are near identical with known-shuffle and unknown-shuffle (41.60\% vs 41.30\%). However, as we increase \textit{k}, the gap between known- and unknown-shuffle becomes increasingly deep (i.e., 34.10\% vs 42.80\% in 20-shot, and 29.95\% vs 40.55\% in 32-shot). Similar effect from increasing \textit{k} is observed on LLaMA-33B and LLaMA-65B. Notably, as \textit{k} increases, the smaller LLaMA-13B also starts experiencing sharp drops with known-shuffle. In the 32-shot setting, its performance drops from 41.29\% (regular 10-shot) to 24.45\%, while remains flat with unknown-shuffle at 40.80\%. For the smallest LLaMA-7B, its performances stay flat across different settings.


The results suggest that the degradation in factual knowledge recall is primarily due to exemplars that contradict models' known knowledge, i.e., counterfactual ICL with known knowledge is essentially teaching LLMs to lie, leading to unexpected results. Additionally, the number of counterfactual exemplars also plays a prominent role. As \textit{k} increases, models experience sharper drops and even smaller models (LLaMA-13B in our experiments) can suffer from significant performance drops. In practical applications, it is therefore important to pair in-context exemplars with the correct answers if known to the model, in order to maximally elicit their parametric knowledge. Finally, we observe comparable performances for known-unshuffle and unknown-unshuffle across different models.

\vspacesection
\section{Fine-tuning}
\vspacesection
\label{sec:Fine-tuning}

In this section, we examine how fine-tuning affects a model's ability to recall factual knowledge and use LLaMA-7B to conduct experiments.

\subsection{Regular fine-tuning}

\textbf{Experimental setup}. We fine-tune LLaMA-7B on the 5k training set and sample 4k additional examples using the same procedure described in Section~\ref{sec:FACT-Bench} as the validation set. We train for 40 steps where training stabilizes based on validation loss and report results on the \Premium subset. 
For model input and output, we use the same input-label format as in the prompting experiments (i.e., input consists of an instruction and a question, and output is the answer to the question).

\textbf{Results}. In the zero-shot setting, we compare models using the \textit{contains} metric instead of EM since the predictions of pretraining-only LLaMA are verbose. Table~\ref{tab:regular fine-tuning} shows that our fine-tuned LLaMA underperforms Vicuna, and both models underperform the pretraining-only LLaMA. Results of this experiment further verify the hypothesis in \citet{zhou2023lima} that a model's knowledge is mostly learned from pretraining, and instruction-tuning only helps align the answer format. 

\begin{table}[htb!]
\begin{center}
\scalebox{0.58}{%
\begin{tabular}{l|rrr}\toprule
\multirow{2}{*}{\bf Models} & \multicolumn{3}{c}{\bf zero-shot}  \\ 
 & {\bf EM} & {\bf F1} & {\bf Contains}  \\\midrule 
LLaMA-7B & 14.65 & 27.66 & 35.20 \\ 
Vicuna-7B & 24.65 & 33.33 & 33.25 \\ 
LLaMA-7B (fine-tuned) & 28.75 & 35.22 & 29.85 \\ 
\bottomrule
\end{tabular}}
\end{center}
\caption{Comparison of LLaMA, Vicuna and our fine-tuned LLaMA.}
\label{tab:regular fine-tuning}
\end{table}


\subsection{Counterfactual fine-tuning}

\textbf{Experimental setup}. In the counterfactual ICL experiments (Section~\ref{sec:The Role of ICL Demonstration}), our experiment results indicate that LLaMA-7B is mostly unaffected by counterfactual exemplars. We set up similar experiments in the fine-tuning setting, where we corrupt the training data with inner-property-shuffle.

\textbf{Results}. Table~\ref{tab:llama-7B-ft-counterfactual} shows the results. Factuality of in-context exemplars plays a critical role for fine-tuning. 
The model can recover part of its capability as training goes on. However, its performance is still significantly worse than that from regular fine-tuning (11.25\% EM vs 29.1\%).

\begin{table}[htb!]
\begin{center}
\scalebox{0.58}{%
\begin{tabular}{l|rrr}\toprule
\multirow{2}{*}{\bf Setup for fine-tuning} & \multicolumn{3}{c}{\bf zero-shot} \\ 
  & {\bf EM} & {\bf F1} & {\bf Contains} \\\midrule 
Regular fine-tuning & 28.75 & 35.22 & 29.85 \\ 
Counterfactual fine-tuning & 10.75 & 15.61 & 12.45 \\
\bottomrule
\end{tabular}}
\end{center}
\caption{Fine-tuning LLaMA-7B with counterfactual knowledge.}
\label{tab:llama-7B-ft-counterfactual}
\end{table}
\vspace{-4mm}

\subsection{Fine-tuning with \textit{known}, \textit{unknown} and \textit{mixed} knowledge}

\textbf{Experimental setup}. 
We fine-tune LLaMA-7B with three types of \textit{factual} knowledge separately: (1) \textit{known}. (2) \textit{unknown}. (3) \textit{mixed}. To approximate known and unknown knowledge, we use the same method described in Section~\ref{subsec:icl counterfactual known unknown}. We use our evaluation set (not including \Premiumns) as the candidate pool to select training data since we need to distinguish between known and unknown knowledge, and 5k is insufficient. We then randomly choose 2.5k training examples for known and unknown knowledge, respectively. 

\textbf{Results}. Table~\ref{tab:fine-tuning known unknown} shows the results. Training with known knowledge consistently outperforms training with mixed knowledge, and training with unknown knowledge leads to the worst performance. The results verify the claim in \citet{schulman-berkeley-talk} that fine-tuning on knowledge unknown to the model teaches the model to hallucinate.


\begin{table}[htb!]
\begin{center}
\scalebox{.58}{%
\begin{tabular}{l|rrr}\toprule
\multirow{2}{*}{\bf Setup for fine-tuning} & \multicolumn{3}{c}{\bf zero-shot}\\ 
 & {\bf EM} & {\bf F1} & {\bf Contains} \\\midrule
Known knowledge & 33.00 & 39.54 & 33.85 \\
Unknown knowledge & 27.55 & 34.10 & 28.75 \\
Mixed knowledge & 29.30 & 36.36 & 30.25 \\

\bottomrule
\end{tabular}}
\end{center}
\caption{Fine-tuning LLaMA-7B with known, unknown and mixed knowledge.}
\label{tab:fine-tuning known unknown}
\end{table}

\vspacesection
\section{Related Work}
\vspacesection
\textbf{Factuality Benchmarks} Question answering datasets, such as Natural Questions \citep{kwiatkowski-etal-2019-natural}, TriviaQA \citep{joshi-etal-2017-triviaqa}, WebQuestions \citep{berant-etal-2013-semantic}, TruthfulQA \citep{lin-etal-2022-truthfulqa} have been used to evaluate factuality of language models. LAMA \citep{petroni2019language,petroni2020how} leverages 4 knowledge sources and converts fact triplets into cloze-style questions. More recent works, such as POPQA \citep{mallen-etal-2023-trust} and KoLA \citep{yu2023kola}, construct benchmarks from Wikidata using templates and cover a limited set of property types and domains. Head-to-Tail \citep{sun2023headtotail} creates their benchmark from DBpedia \citep{Auer2007DBpediaAN} with a focus on evaluating LLMs on knowledge at different popularity levels. Compared to previous benchmarks, \Modelsp is more diverse and representative, covering 134 property types, 20 general domains and 3 answer types. We strictly filter Wikidata triplets and generate valid and specific questions whose answers are grounded in Wikipedia.

\textbf{The role of in-context exemplars}
\citet{min-etal-2022-rethinking} studies the role of in-context exemplars and shows that ground-truth labels are not required for ICL. \citet{yoo-etal-2022-ground} revisits the findings and proposes additional metrics to reveal the importance of ground-truth labels. \citet{wei2023larger} conducts similar experiments and finds that overriding semantic priors is an emergent ability of large models. Our counterfactual ICL experiments corroborate this finding, where large models suffer from significant degradation of knowledge recall. We additionally find that contradicting a model's known knowledge is the primary factor leading to this behavior, along with the number of such exemplars. \citet{pan-etal-2023-context} separates task recognition from task learning in studying how ICL leverages demonstrations, and find that task recognition does not drastically improve with model scaling and more exemplars, while task learning does.

\vspacesection
\section{Conclusion}
\vspacesection
In this paper, we introduce \Model, a comprehensive benchmark that focuses on evaluating factual knowledge of LLMs. We conduct experiments on 31 models from 10 model families and investigate the factors that affect their knowledge recall. We find that instruction-tuning can hurt knowledge recall. In studying the effects of counterfactual in-context exemplars, we highlight the role of known and unknown knowledge. We also conduct fine-tuning experiments, where we highlight the importance of factuality in the training data. We hope that release of our benchmark will be beneficial to the community and help facilitate future research.  

\vspacesection
\section{Limitations}
In this work, we strive to benchmark and analyze as many popular LLMs as resource allows. However, due to the fast pace at which models are released and limited resource, we pick representative and available models at the time of our experimentation. Additionally, distinguishing between model known knowledge and unknown knowledge is an ongoing research topic and in Section~\ref{subsec:icl counterfactual known unknown} and \ref{sec:Fine-tuning}, we check if the model can answer the question correctly as a proxy for model known and unknown knowledge.
\vspacesection

\bibliography{iclr2024_conference}
\bibliographystyle{iclr2024_conference}

\appendix

\section{Task Instructions}

Table \ref{appendix:question-generation}, \ref{appendix:zero-shot}, \ref{appendix:10-shot}, and \ref{appendix:counterfactual-10-shot} show the prompts for question generation, zero-shot, 10-shot, and counterfactual 10-shot, respectively.

\begin{table*}
\begin{tcolorbox}[colback=gray!20!white, colframe=gray!60!black, rounded corners]
\footnotesize
Instruction: given a triplet in the form of (subject, property, object), generate a question about the subject.\\

Requirement:

1. The object must be the unique answer to the question!!

2. The question must be specific regarding the nature or category of the object so that the object is the only answer!! For example, if the answer is a city, ask for city; If the answer is city and state, ask for city and state, and so on.

3. The question should not include the object!!

4. If the property is height or elevation above sea level, ask for meters. If the property is area, ask for square kilometers. If the property is length or width, ask for kilometers. If the answer is temperature, ask for celsius.

5. Your response should strictly follow the format:\\

Question:$<$question$>$ Answer: $<$answer$>$.

Triplet: {triplet}
\end{tcolorbox}
\caption{Prompt for question generation.}
\label{appendix:question-generation}
\end{table*}





\begin{table*}
\begin{tcolorbox}[colback=gray!20!white, colframe=gray!60!black, rounded corners]
\footnotesize
Instruction: answer the following question. Don't include explanation. Keep the answer as concise as possible.

Question: \{QUESTION\}

Answer:
\end{tcolorbox}
\caption{Prompt for zero-shot.}    
\label{appendix:zero-shot}
\end{table*}

\begin{table*}
\begin{tcolorbox}[colback=gray!20!white, colframe=gray!60!black, rounded corners]
\footnotesize
Instruction: answer the following question. Don't include explanation. Keep the answer as concise as possible.

Question: In which military branch did Henry Curtis serve?

Answer: Royal Navy \\

Question: Which Australian rules football club was Simon Madden a member of?

Answer: Essendon Football Club\\

Question: What architectural style is Pine Bloom Plantation?

Answer: Greek Revival architecture\\

Question: On what date was Ed Hooper born?

Answer: March 10, 1964\\

Question: Who founded Tangerine Dream?

Answer: Edgar Froese\\

Question: Who is the performer of 'Hollywood's Not America'?

Answer: Ferras\\

Question: What company manufactured the AMC Gremlin?

Answer: American Motors Corporation\\

Question: In what year was the Whistler House Museum of Art officially opened?

Answer: 1908\\

Question: What is Ellen S. Baker's mother's name?

Answer: Claire Shulman\\

Question: How many floors above ground does Premier Tower have?

Answer: 78\\

Question: \{QUESTION\}

Answer:
\end{tcolorbox}
\caption{Prompt for few-shot (10-shot).}
\label{appendix:10-shot}
\end{table*}

\begin{table*}
\begin{tcolorbox}[colback=gray!20!white, colframe=gray!60!black, rounded corners]
\footnotesize
Instruction: answer the following question. Don't include explanation. Keep the answer as concise as possible.

Question: In which military branch did Henry Curtis serve?

Answer: United States Marine Corps \\

Question: Which Australian rules football club was Simon Madden a member of?

Answer: Tennessee Volunteers football \\

Question: What architectural style is Pine Bloom Plantation?

Answer: Art Nouveau \\

Question: On what date was Ed Hooper born?

Answer: October 12, 1876 \\

Question: Who founded Tangerine Dream?

Answer: Frank Varga \\

Question: Who is the performer of 'Hollywood's Not America'?

Answer: Damien Bodie \\

Question: What company manufactured the AMC Gremlin?

Answer: Kalem Company \\

Question: In what year was the Whistler House Museum of Art officially opened?

Answer: 1832 \\

Question: What is Ellen S. Baker's mother's name?

Answer: Empress Dayi \\

Question: How many floors above ground does Premier Tower have?

Answer: 164 \\

Question: \{QUESTION\}

Answer:
\end{tcolorbox}
\caption{Prompt for counterfactual few-Shot (10-shot).}
\label{appendix:counterfactual-10-shot}
\end{table*}

\section{Experimental Settings}
\subsection{Zero-Shot and Few-Shot ICL Experiments}
For open models, we use Python version 3.10, Torch version 2.0.0, and the Hugging Face Transformers library \citep{wolf-etal-2020-transformers} with version 4.31.0. We use greedy decoding for reproducibility. Batch size is set to 4 and sequences are left-padded with [PAD] token set to [EOS] token if it's not already set. All our experiments were conducted on A100 GPUs with 40GB of RAM. 

\subsection{Fine-tuning Experiments}
For fine-tuning LLaMA-7B, we use the AdamW optimizer with a learning rate of 0.0001 and a cosine learning rate scheduler. Effective batch size is set to 512 and sequence length to 256. We train for 40 steps, where we observe the training stabilizes based on validation set performance.




\subsection{Full Benchmarking Results on \Premiumns}
\label{subsec:full premium2k}
Table~\ref{table: full results} shows zero-shot, 1-shot, 6-shot, and 10-shot results on \Premium for all 31 models we consider.

\begin{table*}[htb!]
\begin{center}
\scalebox{0.55}{%
\begin{tabular}{l|rrr|rrr|rrr|rrr}\toprule
\multirow{2}{*}{\bf Models} & \multicolumn{3}{c|}{\bf zero-shot} & \multicolumn{3}{c|}{\bf 1-shot} & \multicolumn{3}{c|}{\bf 6-shot} & \multicolumn{3}{c}{\bf 10-shot} \\ & {\bf EM} & {\bf F1} & \rotatebox[origin=c]{90}{\bf Contains} & {\bf EM} & {\bf F1} & \rotatebox[origin=c]{90}{\bf Contains} & {\bf EM} & {\bf F1} & \rotatebox[origin=c]{90}{\bf Contains} & {\bf EM} & {\bf F1} & \rotatebox[origin=c]{90}{\bf Contains} \\\midrule
GPT-4 & 58.60 & 65.99 & 64.65 & 59.85 & 65.7 & 63.2 & 63.35	& 69.17	& 66.45 & 65.90 & 72.01 & 69.15 \\
GPT-3.5-turbo & 49.75 & 55.77 & 52.60 & 51.25 & 57.41 & 53.70 & 52.65 & 59.36 & 55.80 & 53.55 & 60.19 & 56.40 \\\midrule
BLOOM-7.1B & 3.20 & 14.31 & 20.30 & 18.95 & 24.90 & 19.95 & 17.85 & 24.98 & 19.90 & 18.15 & 24.76 & 19.75 \\
BLOOMZ-7.1B & 18.00 & 24.08 & 19.45 & 14.05 & 18.02 & 15.35 & 14.40 & 19.65 & 17.05 & 15.20 & 20.49 & 17.70 \\
LLaMA-7B & 14.65 & 27.66 & 35.20 & 33.25 & 39.73 & 34.15 & 35.55 & 42.88 & 37.15 & 35.05 & 41.96 & 36.75 \\
LLaMA-13B & 21.35 & 33.92 & 39.95 & 36.45 & 42.81 & 37.30 & 41.15 & 47.98 & 42.75 & 41.20 & 48.48 & 42.95 \\
LLaMA-33B & 27.25 & 40.48 & 46.55 & 45.25 & 51.70 & 46.70 & 48.30 & 55.03 & 50.30 & 48.90 & 55.65 & 51.10 \\
LLaMA-65B & 35.25 & 46.67 & 49.20 & 47.15 & 53.33 & 48.45 & 52.15 & 58.44 & 53.80 & 52.45 & 58.98 & 54.10 \\
Vicuna-7B-v1.3 & 24.65 & 33.33 & 33.25 & 31.15 & 38.30 & 33.80 & 30.10 & 39.17 & 35.05 & 31.00 & 39.23 & 34.65 \\
Vicuna-13B-v1.3 & 32.95 & 39.74 & 35.15 & 36.45 & 42.63 & 37.60 & 38.00 & 46.14 & 41.20 & 38.40 & 45.79 & 41.15 \\
Vicuna-33B-v1.3 & 34.30 & 43.94 & 44.15 & 41.39 & 49.38 & 44.75 & 44.10 & 52.36 & 48.10 & 44.00 & 52.44 & 48.05 \\
OpenLLaMA-7B & 14.05 & 26.62 & 32.30 & 31.75 & 38.03 & 32.80 & 32.55 & 40.19 & 34.70 & 33.80 & 41.23 & 35.95 \\
OpenLLaMA-13B & 25.70 & 35.98 & 37.35 & 37.05 & 43.21 & 38.40 & 38.75 & 45.98 & 40.70 & 39.70 & 47.07 & 41.55 \\
FLAN-T5-XXL (11B) & 20.60 & 27.63 & 21.60 & 20.45 & 27.31 & 21.45 & 21.05 & 27.93 & 22.15 & 20.95 & 27.87 & 22.00 \\
T0++ (11B) & 16.05 & 24.46 & 21.25 & 16.75 & 23.40 & 19.95 & 16.80 & 24.43 & 20.00 & 17.05 & 24.64 & 19.85 \\
UL2 (20B) & 3.40 & 13.83 & 23.55 & 23.50 & 28.91 & 24.40 & 24.15 & 30.92 & 25.75 & 23.50 & 30.41 & 25.00 \\
FLAN-UL2 (20B) & 24.05 & 30.58 & 25.20 & 24.10 & 30.64 & 25.25 & 24.10 & 30.47 & 25.30 & 23.90 & 30.26 & 24.95 \\
Falcon-7B & 23.60 & 31.03 & 30.05 & 30.25 & 36.35 & 31.90 & 30.70 & 37.19 & 32.60 & 30.45 & 37.19 & 32.25 \\
Falcon-7B-instruct & 10.85 & 20.59 & 25.10 & 21.75 & 29.25 & 24.60 & 22.45 & 30.51 & 25.45 & 22.45 & 30.10 & 25.20 \\
Falcon-40B & 26.55 & 31.81 & 30.90 & 39.10 & 45.68 & 40.50 & 42.05 & 48.70 & 43.60 & 42.25 & 48.95 & 43.80 \\
Falcon-40B-instruct & 21.95 & 32.98 & 40.25 & 38.85	& 45.42 & 40.75 & 40.40 & 47.04 & 42.20 & 40.00 & 47.14 & 41.85 \\
Falcon-180B & 44.90 & 50.34 & 47.45 & 49.25 & 55.80 & 50.60 & 53.55 & 59.89 & 55.05 & 53.45 & 59.94 & 55.00 \\
Falcon-180B-chat & 39.95 & 48.14 & 47.10 & 47.00 & 54.14 & 49.30 & 49.05 & 56.08 & 51.50 & 49.30 & 56.17 & 51.60 \\
MPT-7B & 3.45 & 14.89 & 30.35 & 28.85 & 34.72 & 29.85 & 29.75 & 36.00 & 31.15 & 30.45 & 36.51 & 31.55 \\
MPT-7B-instruct & 3.55 & 11.34 & 30.40 & 21.55 & 27.84 & 29.25 & 26.35 & 32.95 & 29.30 & 27.85 & 34.31 & 29.60 \\
MPT-30B & 25.30 & 34.18 & 35.00 & 34.35	& 40.73 & 35.55 & 35.80 & 42.98 & 37.55 & 36.05 & 43.03 & 37.75 \\
MPT-30B-instruct & 19.05 & 28.41 & 33.50 & 28.80 & 35.70 & 31.20 & 31.00 & 38.06 & 33.50 & 31.50 & 38.30 & 33.85 \\
Pythia-6.9B & 11.00 & 14.11 & 13.15 & 21.20 & 27.06 & 22.45 & 21.85 & 28.41 & 23.05 & 21.70 & 28.32 & 23.25 \\
Pythia-12B & 15.25 & 22.98 & 22.00 & 22.75 & 28.71 & 23.70 & 22.95 & 29.44 & 24.35 & 23.20 & 29.73 & 24.65 \\
Mistral-7B & 28.45 & 31.34 & 29.25 & 38.90 & 45.38 & 39.8 & 40.45 & 47.43 & 41.85 & 40.75 & 47.96 & 42.60 \\
Mistral-7B-instruct & 26.00 & 32.97 & 29.30 & 26.80 & 34.68 & 30.05 & 26.80 & 35.06 & 30.35 & 27.20 & 35.43 & 30.75 \\
\bottomrule
\end{tabular}%
}%
\end{center}
\caption{Full benchmarking results on \Premiumns.}
\label{table: full results}
\end{table*}

\subsection{Full Benchmarking Results on 15k Evaluation Set}
\label{subsec:15k eval}
Table~\ref{tab:15k-eval} shows zero-shot and 10-shot results on the full 15k evaluation set for all 31 models we consider.

\begin{table*}[htb]
\centering
\scalebox{0.7}{%
\begin{tabular}{l|rrr|rrr}
\toprule
\multirow{2}{*}{\bf Models} & \multicolumn{3}{c|}{\bf Zero-shot} & \multicolumn{3}{c}{\bf 10-shot}\\
 & {\bf EM} & {\bf F1} & {\bf Contains} & {\bf EM} & {\bf F1} & {\bf Contains} \\
\midrule
GPT-4 & 52.31 & 59.83 & 58.47 & 58.30 & 64.72 & 61.39 \\
GPT-3.5-turbo & 44.36 & 50.21 & 47.19 & 48.60 & 55.21 & 51.55 \\
BLOOM-7.1B & 3.37 & 14.09 & 18.87 & 17.05 & 23.58 & 18.57 \\
BLOOMZ-7.1B & 16.54 & 22.78 & 17.96 & 14.57 & 20.10 & 17.01 \\
LLaMA-7B & 12.53 & 25.16 & 32.69 & 32.24 & 39.16 & 33.99 \\
LLaMA-13B & 18.53 & 31.25 & 36.79 & 37.33 & 44.30 & 39.16 \\
LLaMA-33B & 24.94 & 37.51 & 43.29 & 44.31 & 51.05 & 46.21 \\
LLaMA-65B & 31.19 & 42.55 & 45.62 & 47.39 & 53.67 & 49.02 \\
Vicuna-7B-v1.3 & 22.46 & 31.07 & 31.09 & 28.14 & 36.72 & 32.18 \\
Vicuna-13B-v1.3 & 29.81 & 36.17 & 32.42 & 34.21 & 42.09 & 37.45 \\
Vicuna-33B-v1.3 & 30.23 & 40.07 & 40.69 & 38.69 & 47.06 & 42.97 \\
OpenLLaMA-7B & 12.06 & 24.47 & 30.79 & 31.11 & 38.28 & 33.12 \\
OpenLLaMA-13B & 23.06 & 33.31 & 34.45 & 35.19 & 42.21 & 37.03 \\
FLAN-T5-XXL & 19.07 & 25.88 & 20.42 & 19.32 & 26.28 & 20.73 \\
T0++ & 14.59 & 22.97 & 19.45 & 15.39 & 23.54 & 18.47 \\
UL2 & 3.37 & 13.48 & 22.05 & 21.85 & 28.72 & 23.29 \\
FLAN-UL2 & 21.16 & 27.97 & 22.62 & 21.03 & 27.86 & 22.53 \\
Falcon-7B & 21.29 & 28.84 & 27.50 & 28.00 & 34.95 & 29.75 \\
Falcon-7B-instruct & 9.50 & 19.14 & 23.15 & 19.62 & 27.34 & 22.81 \\
Falcon-40B & 23.41 & 28.56 & 28.57 & 37.82 & 44.64 & 39.37 \\
Falcon-40B-instruct & 18.79 & 29.79 & 36.87 & 36.53 & 43.14 & 38.32 \\
MPT-7B & 2.95 & 14.23 & 28.04 & 27.85 & 34.39 & 29.13 \\
MPT-7B-instruct & 3.72 & 11.31 & 28.39 & 25.89 & 32.55 & 27.61 \\
MPT-30B & 22.69 & 31.58 & 31.87 & 33.59 & 40.21 & 35.15 \\
MPT-30B-instruct & 16.51 & 26.01 & 30.29 & 29.37 & 36.35 & 31.53 \\
Pythia-6.9B & 9.26 & 12.55 & 11.93 & 19.83 & 26.39 & 21.43 \\
Pythia-12B & 13.81 & 21.48 & 20.19 & 21.40 & 28.17 & 22.88 \\
\bottomrule
\end{tabular}%
}
\caption{Full benchmarking results on 15K evaluation set.}
\label{tab:15k-eval}
\end{table*}

\section{List of Properties}
\label{sec:all properties}
Table \ref{tab:date-properties}, \ref{tab:number-properties}, and \ref{tab:entity-properties} show the 134 property types
by answer type (date, number, and entity).

\begin{table*}
\centering
\resizebox{\textwidth}{!}{%
\begin{tcolorbox}[colback=white,colframe=gray!60!black]
\footnotesize
\begin{multicols}{3}
    \begin{itemize}
    \item date of birth (P569)
    \item date of death (P570)
    \item inception (P571)
    \item time of discovery or invention (P575)
    \item publication date (P577)
    \item first flight (P606)
    \item UTC date of spacecraft launch (P619)
    \item UTC date of spacecraft landing (P620)
    \item date of disappearance (P746)
    \item date of first performance (P1191)
    \item date of official opening (P1619)
    \item production date (P2754)
    \item date of official closure (P3999)
    \item recording date (P10135)
    \item dissolved, abolished or demolished date (P576)
    \item start time (P580)
    \item end time (P582)
    \item service entry (P729)
    \item service retirement (P730)
    \item discontinued date (P2669)
    \item debut date (P10673)*
    \item date of incorporation (P10786)*
    \end{itemize}
\end{multicols}%
\end{tcolorbox}
}
\caption{List of 22 date properties.  The ones with asterisk do not appear in \Premiumns.}
\label{tab:date-properties}
\end{table*}

\begin{table*}
\centering
\resizebox{\textwidth}{!}{%
\begin{tcolorbox}[colback=white,colframe=gray!60!black]
\footnotesize
\begin{multicols}{3}
    \begin{itemize} 
    \item maximum capacity (P1083)
    \item atomic number (P1086)  
    \item total produced (P1092)
    \item number of cylinders (P1100)
    \item floors above ground (P1101)
    \item number of platform tracks (P1103)
    \item number of episodes (P1113)
    \item number of deaths (P1120)
    \item neutron number (P1148)
    \item minimum number of players (P1872)
    \item maximum number of players (P1873)
    \item number of children (P1971)
    \item length (P2043)
    \item elevation above sea level (P2044) 
    \item area (P2046)
    \item height (P2048)
    \item width (P2049)
    \item mass (P2067)
    \item melting point (P2101)
    \item chromosome count (P5230)
    \item number of seats in legislature (P1410)*
    \item memory capacity (P2928)*
    \end{itemize}
\end{multicols}%
\end{tcolorbox}
}
\caption{List of 22 number properties.  The ones with asterisk do not appear in \Premiumns.}
\label{tab:number-properties}
\end{table*}
\begin{table*}
\centering
\resizebox{\textwidth}{!}{%
\begin{tcolorbox}[colback=white,colframe=gray!60!black]
\footnotesize
\begin{multicols}{3}
    \begin{itemize}
    \item member of political party (P102)
    \item taxon rank (P105)
    \item occupation (P106)  
    \item location of creation (P1071)
    \item founded by (P112)
    \item airline hub (P113)
    \item home venue (P115)  
    \item league (P118)
    \item place of burial (P119)
    \item publisher (P123)
    \item owned by (P127)
    \item located in the administrative territorial entity (P131)  
    \item participant in (P1344)
    \item winner (P1346)
    \item movement (P135)
    \item genre (P136)
    \item operator (P137)
    \item capital of (P1376)
    \item licensed to broadcast to (P1408)
    \item IUCN conservation status (P141) 
    \item languages spoken, written or signed (P1412)
    \item affiliation (P1416)
    \item present in work (P1441)
    \item architectural style (P149)
    \item country for sport (P1532)
    \item headquarters location (P159)
    \item transport network (P16)
    \item producer (P162)
    \item award received (P166)
    \item country (P17)
    \item creator (P170)
    \item parent taxon (P171)
    \item performer (P175)
    \item manufacturer (P176)
    \item crosses (P177)
    \item developer (P178)
    \item endemic to (P183)
    \item doctoral advisor (P184)
    \item place of birth (P19)  
    \item collection (P195)
    \item place of death (P20)
    \item cuisine (P2012)
    \item basin country (P205)
    \item located in or next to body of water (P206) 
    \item father (P22)
    \item military branch (P241)
    \item mother (P25)
    \item record label (P264)
    \item country of citizenship (P27)
    \item production company (P272)
    \item location (P276)
    \item programmed in (P277)
    \item designed by (P287)
    \item vessel class (P289)
    \item continent (P30)
    \item operating system (P306)
    \item capital (P36)
    \item part of (P361)
    \item space launch vehicle (P375)
    \item parent astronomical body (P397)
    \item mouth of the watercourse (P403)
    \item position played on team / speciality (P413)  
    \item original broadcaster (P449)
    \item color (P462) 
    \item occupant (P466)
    \item animal breed (P4743)
    \item court (P4884)
    \item country of origin (P495)
    \item author (P50)
    \item cause of death (P509)
    \item school district (P5353)
    \item member of sports team (P54)
    \item director (P57)
    \item screenwriter (P58)
    \item conflict (P607)
    \item discoverer or inventor (P61)
    \item highest point (P610) 
    \item sport (P641)
    \item drafted by (P647)
    \item educated at (P69)  
    \item diocese (P708)
    \item location of formation (P740)  
    \item parent organization (P749)
    \item distributed by (P750)
    \item historic county (P7959)
    \item country of registry (P8047)
    \item architect (P84)
    \item composer (P86)
    \item filming location (P915) 
    \item allegiance (P945)
    \end{itemize}
\end{multicols}%
\end{tcolorbox}
}
\caption{List of 90 entity properties.}
\label{tab:entity-properties}
\end{table*}


\end{document}